\documentclass[conference]{IEEEtran}
\IEEEoverridecommandlockouts

\usepackage{booktabs}
\usepackage{multirow}
\usepackage{hyperref}
\usepackage{enumitem}
\makeatletter

\def\ps@IEEEtitlepagestyle{%
  \def\@oddfoot{\mycopyrightnotice}%
  \def\@evenfoot{}%
}
\def\mycopyrightnotice{%
  {\footnotesize 979-8-3315-3562-9/25/\$31.00~\copyright~2025 IEEE\hfill}%
  \gdef\mycopyrightnotice{}%
}

\usepackage{blindtext}
\usepackage{eso-pic}
\IEEEoverridecommandlockouts
\usepackage{cite}
\usepackage{amsmath,amssymb,amsfonts}
\usepackage{algorithmic}
\usepackage{graphicx}
\usepackage{textcomp}
\usepackage{xspace}
\newcommand{\BfPara}[1]{\vspace{0.1em}{\noindent\bf#1.}\xspace}
\usepackage{xcolor}
\def\BibTeX{{\rm B\kern-.05em{\sc i\kern-.025em b}\kern-.08em
    T\kern-.1667em\lower.7ex\hbox{E}\kern-.125emX}}
    
\usepackage{eso-pic}
\newcommand\AtPageUpperMyright[1]{\AtPageUpperLeft{%
 \put(\LenToUnit{0.17\paperwidth},\LenToUnit{-2cm}){%
     \parbox{0.9\textwidth}{\raggedleft\fontsize{8}{11}\selectfont #1}}%
 }}%
\newcommand{\conf}[1]{%
\AddToShipoutPictureBG*{%
\AtPageUpperMyright{#1}
}
}    

\makeatletter
\newcommand{\linebreakand}{%
  \end{@IEEEauthorhalign}
  \hfill\mbox{}\par
  \mbox{}\hfill\begin{@IEEEauthorhalign}
}
\makeatother

    \usepackage{balance}

\begin{document}
\title{\vspace*{1cm} Uncertainty-Guided Coarse-to-Fine Tumor Segmentation with Anatomy-Aware Post-Processing\\
%{\footnotesize \textsuperscript{*}Note: Sub-titles are not captured in Xplore and should not be used}
\thanks{Dr. Bagci acknowledges the following grants: NIH R01-CA240639, and FDOH (Florida Department of Health) through the James and Esther King Biomedical Research Program-20K04. Code and pipelines are available at: \url{https://github.com/ilkinisler/CoarseToFineSegmentation}.}
}

\author{\IEEEauthorblockN{Ilkin Sevgi Isler}
\IEEEauthorblockA{\textit{Department of Computer Science}\\
\textit{University of Central Florida}\\
Orlando, FL, USA \\
ilkin@ucf.edu}
\and

\IEEEauthorblockN{David Mohaisen}
\IEEEauthorblockA{\textit{Department of Computer Science} \\
\textit{University of Central Florida}\\
Orlando, FL, USA \\
mohaisen@ucf.edu}
\and
\IEEEauthorblockN{Curtis Lisle}
\IEEEauthorblockA{\textit{KnowledgeVis} \\
\textit{LLC}\\
Altamonte Springs, FL, USA \\
clisle@knowledgevis.com}
\linebreakand
\IEEEauthorblockN{Damla Turgut}
\IEEEauthorblockA{\textit{Department of Computer Science} \\
\textit{University of Central Florida}\\
Orlando, FL, USA \\
damla.turgut@ucf.edu}
\and
\IEEEauthorblockN{Ulas Bagci}
\IEEEauthorblockA{\textit{Department of Radiology} \\
\textit{Northwestern University}\\
Chicago, IL, USA \\
ulas.bagci@northwestern.edu}
}

\maketitle
\conf{\textit{Proc. of International Conference on Artificial Intelligence, Computer, Data Sciences and Applications (ACDSA 2025) \\ 
7-9 August 2025, Antalya-Türkiye}}

\begin{abstract}
Reliable tumor segmentation in thoracic computed tomography (CT) remains challenging due to boundary ambiguity, class imbalance, and anatomical variability. 
We propose an uncertainty-guided, coarse-to-fine segmentation framework that combines full-volume tumor localization with refined region-of-interest (ROI) segmentation, enhanced by anatomically aware post-processing. 
The first-stage model generates a coarse prediction, followed by anatomically informed filtering based on lung overlap, proximity to lung surfaces, and component size. 
The resulting ROIs are segmented by a second-stage model trained with uncertainty-aware loss functions to improve accuracy and boundary calibration in ambiguous regions. 
Experiments on private and public datasets demonstrate improvements in Dice and Hausdorff scores, with fewer false positives and enhanced spatial interpretability. 
These results highlight the value of combining uncertainty modeling and anatomical priors in cascaded segmentation pipelines for robust and clinically meaningful tumor delineation. 
On the Orlando dataset, our framework improved Swin UNETR Dice from 0.4690 to 0.6447. 
Reduction in spurious components was strongly correlated with segmentation gains (HD95: $\rho = -0.83$, $p < 0.0001$), underscoring the value of anatomically informed post-processing.

\end{abstract}

%\copyrightnotice{XXX-X-XXXX-XXXX-X/XX/\$XX.00 ©20XX IEEE}


\section{Introduction}
The accurate and reliable segmentation of lung tumors in computed tomography (CT) images remains a persistent challenge in thoracic oncology and medical image analysis~\cite{armato2011lung}. Performance often suffers due to tumor heterogeneity, class imbalance, partial volume effects, and boundary ambiguity caused by both intra- and inter-patient variation~\cite{zhou2019lung}. These issues are particularly acute near the pleura, mediastinum, and central hilar regions, where tumors may cross anatomical boundaries or exhibit indistinct contours~\cite{tong2020fully}. While most gross tumor volumes (GTVs) reside within the lung parenchyma, a clinically important subset partially extends beyond the lung walls~\cite{tang2022lung}. This makes it difficult to rely on rigid anatomical priors such as lung-only masking~\cite{isensee2021nnu}.

Despite advances in deep learning-based segmentation, few models transition into clinical workflows. A key barrier is the lack of interpretability and error awareness in deterministic systems. Clinical integration requires more than high Dice scores; it demands spatial reliability, manageable false positives, and transparency in uncertain regions~\cite{kompa2021second}. This highlights the need for segmentation frameworks that incorporate uncertainty modeling and anatomical reasoning not as optional features, but as essential elements for clinical deployment~\cite{hesamian2019deep}.

%3. The paper uses the term "end-to-end" to describe the framework. While the stages are connected to form a complete pipeline, the common understanding of "end-to-end trainable" implies joint optimization of all learnable components. Given the separate training of Stage 1 and Stage 2 models and the rule-based post-processing, "fully automated pipeline" might be more precise terminology.
\BfPara{Our Contribution and Findings} To address these needs, we propose a two-stage, fully automated segmentation framework that combines coarse tumor localization with refined region-of-interest (ROI) segmentation. In the first stage, a full-volume model identifies candidate tumor regions. These are refined through a clinically informed post-processing pipeline that applies anatomical constraints, including lung overlap thresholds, minimum surface distance, and region-specific filters. Rather than introducing an explicit localization model, we treat segmentation as a surrogate for localization. We show that lightweight, anatomy-aware post-processing can yield high spatial fidelity without increasing computational load.

We also show that reducing the number of candidate components passed to the second-stage model improves both segmentation accuracy and predictive confidence. This supports the view that component selection functions as a form of spatial attention. Focusing on the most salient regions enhances robustness, especially in challenging cases. These improvements are achieved without added architectural complexity, underscoring the effectiveness of domain-informed filtering over brute-force model scaling.

In the second stage, the model incorporates uncertainty-aware loss functions to adaptively optimize predictions in high-ambiguity areas, particularly at tumor boundaries. This uncertainty modeling enhances spatial calibration and provides interpretable outputs, offering clinicians an additional layer of review that can strengthen confidence in the results.

We evaluate the framework through a series of experiments that examine the influence of ROI context, component count, and uncertainty-driven training objectives. The results demonstrate that the proposed anatomy-aware, uncertainty-refined segmentation system achieves competitive accuracy while improving interpretability and clinical plausibility. These findings advance the practical utility of automated tumor segmentation in thoracic imaging.

\begin{figure}[t]
    \centering
    \includegraphics[width=\columnwidth]{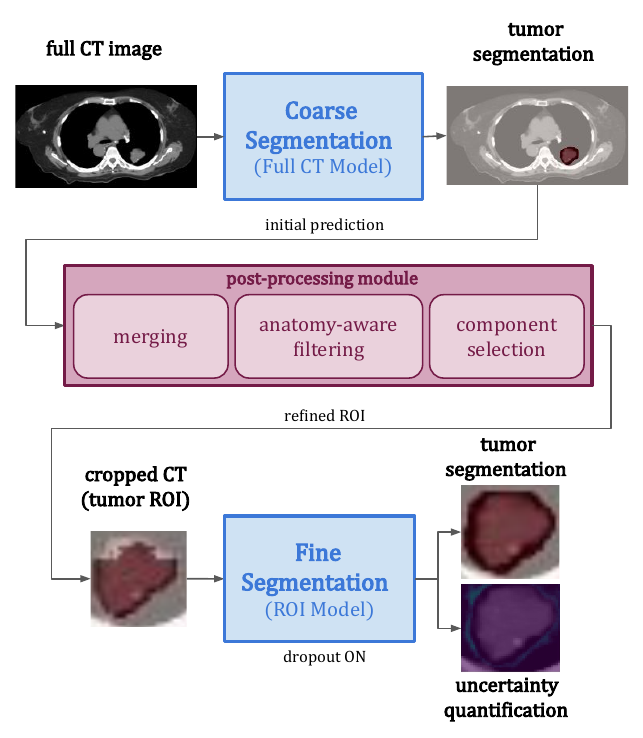}\vspace{-3mm}
    \caption{Overview of the proposed two-stage tumor segmentation framework. The first-stage model performs coarse segmentation on the full CT scan and outputs both a tumor prediction and an uncertainty map. Post-processing module uses these outputs to generate a refined input region. The second-stage model conducts fine segmentation on the cropped ROI to produce the final tumor mask. The uncertainty map enhances interpretability and supports adaptive refinement around ambiguous boundaries.}
    \label{fig:main}\vspace{-5mm}
\end{figure}

\begin{figure}[t]
    \centering
    \includegraphics[width=\linewidth]{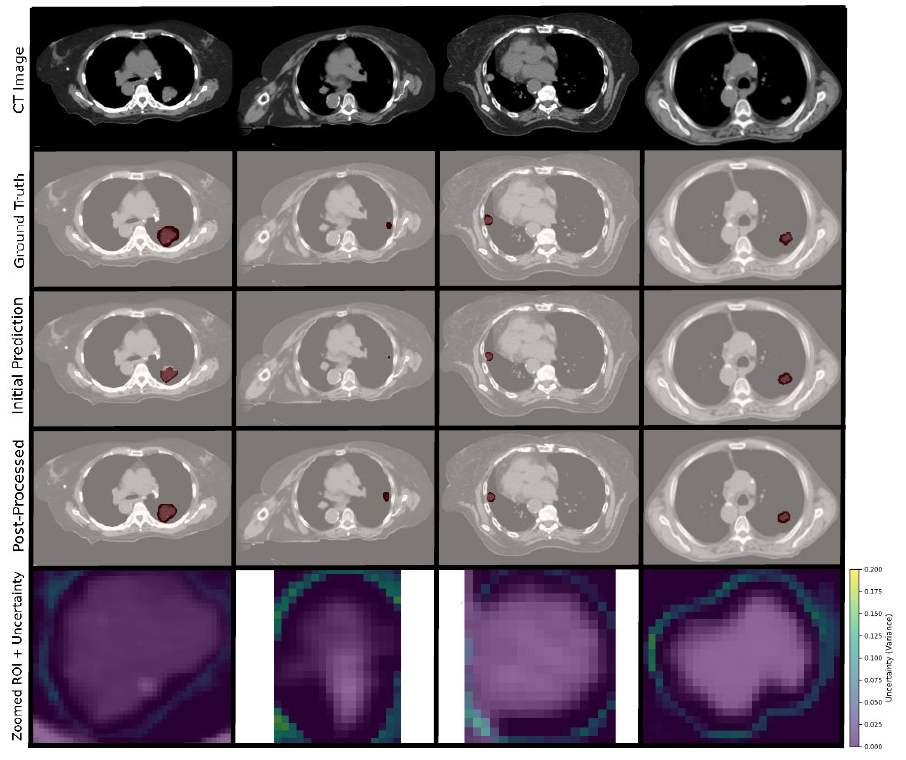}
    \caption{Example patient cases from our two-stage segmentation pipeline. From left to right: full CT image, ground truth mask, initial prediction, post-processed prediction, and the region-of-interest prediction with uncertainty map overlay. Our pipeline demonstrates improved tumor delineation and spatial accuracy through uncertainty modeling and anatomical post-processing.}
    \label{fig:example_patients}\vspace{-3mm}
\end{figure}

\BfPara{Organization} The rest of this paper is organized as follows. In section~\ref{sec:related} we review the related work. In section~\ref{sec:method}, we present our methodology, including dataset and preprocessing, our two-stage segmentation framework, component selection strategy, and uncertainty-aware loss function with model architecture and post-processing. In section~\ref{sec:results} we present our results and analysis, followed bb conclusion in section~\ref{sec:conclusion}

\section{Related Work}\label{sec:related}
There have been several studies related to this work, focusing on lung tumor segmentation, region-of-interest learning, uncertainty-aware segmentation, anatomically informed post-processing, and considerations for clinical translation, which we review in the following.

\BfPara{Lung Tumor Segmentation}  
Deep learning-based methods have significantly advanced lung tumor segmentation in CT, with architectures such as U-Net~\cite{ronneberger2015u} and its 3D extensions becoming foundational, producing state-of-the-art results. Variants like UNETR~\cite{hatamizadeh2022unetr} and Swin UNETR~\cite{tang2022self} leverage transformer-based encoders for better global context understanding. However, segmenting tumors that are small, irregular, or located near anatomical boundaries remains challenging due to class imbalance and limited spatial cues.

\BfPara{Region-of-Interest (ROI) Learning}  
Cascade segmentation approaches decompose the problem into coarse localization followed by fine segmentation, often improving accuracy while reducing computational cost~\cite{isensee2021nnu}. Prior work has explored tight cropping~\cite{li2018h} or context-aware padding~\cite{zhou2019cascaded} to balance precision and spatial understanding. Our work adopts a similar strategy but incorporates domain-specific post-processing to enhance ROI selection reliability.

\BfPara{Uncertainty-Aware Segmentation}
Uncertainty modeling has emerged as a powerful tool for enhancing model interpretability in medical imaging with multiple competing results~\cite{kendall2017uncertainties}. Techniques such as Monte Carlo dropout~\cite{gal2016dropout}, test-time augmentation~\cite{wang2019aleatoric}, and ensemble methods~\cite{mehrtash2020confidence} allow quantification of predictive variance. These approaches support adaptive learning and improve robustness, particularly in ambiguous regions. Recent work introduced MC-Swin-U~\cite{isler2023self}, a self-supervised transformer model that integrates Monte Carlo dropout for voxel-wise uncertainty estimation in lung tumor and organ-at-risk segmentation. We build upon this foundation by integrating uncertainty into the ROI model’s loss function to focus optimization on high-ambiguity areas.

\BfPara{Anatomically Informed Post-Processing}  
Several studies apply anatomical priors to filter out implausible predictions. For example, lung masking~\cite{hofmanninger2020automatic} and anatomical constraints based on spatial location or proximity to lung surfaces have been used to reduce false positives~\cite{yang2017automatic}. Our method extends this idea through a multi-criteria filtering scheme that incorporates lung overlap thresholds, distance-based heuristics, and component size criteria to ensure that only clinically plausible regions are refined in the second stage.

\BfPara{Clinical Translation Considerations}  
Despite high Dice scores in the experimental settings, many segmentation systems fail in clinical deployment due to the lack of interpretability and error awareness~\cite{litjens2017survey}. As such, recent trends emphasized the need for pipelines that offer spatial reliability, explainability, and robustness to distribution shift~\cite{reinke2021common}. To this end, our approach addresses these needs by combining spatial filtering, ROI segmentation, and uncertainty quantification in a modular, interpretable framework.

\section{Methodology}\label{sec:method}

This study proposes a two-stage, fully automated segmentation framework that integrates coarse tumor localization, uncertainty-aware ROI refinement, and anatomically guided post-processing. The goal is to enhance segmentation precision, robustness, and interpretability in thoracic CT scans while maintaining computational efficiency. The methodology is structured to evaluate segmentation performance under variations in ROI context, component selection, uncertainty modeling, model architecture, and dataset generalization.

\subsection{Datasets and Preprocessing}

This study utilizes two datasets for training and evaluation. The primary dataset consists of thoracic CT scans collected from a private institutional source, with expert-annotated tumor segmentation. Due to institutional privacy restrictions, this dataset is not publicly available but serves as the foundation for model development and internal validation. To assess generalization, the proposed pipeline was replicated on the publicly available NSCLC-Radiomics dataset using the same experimental configuration, enabling reproducibility and comparative analysis across domains.

All CT scans were resampled to an isotropic resolution and intensity-normalized to a fixed window prior to model input. Lung masks were generated using a pre-trained lung segmentation model and applied during post-processing.

The same pre-processing, ROI extraction, and component selection procedures were used for both datasets to ensure consistency. This unified pipeline enables the evaluation of segmentation accuracy, model robustness, and uncertainty calibration under real-world variability and cross-institutional generalizability, which are ideal for systemic evaluation.

\subsection{Two-Stage (Cascade) Segmentation Framework}

The proposed pipeline follows a coarse-to-fine, cascade segmentation architecture designed to decompose the tumor segmentation task into two sub-problems: (1) coarse localization and (2) boundary refinement.

\noindent \textbf{Stage 1 (Coarse Localization):} The full-resolution CT volume is processed by a 3D segmentation model to generate a coarse tumor prediction map. This stage operates deterministically, without dropout or uncertainty modeling. Connected component analysis is then applied to the binary prediction to extract candidate regions for further refinement.

\noindent \textbf{Stage 2 (ROI-Based Refinement):} A configurable number of top-$k$ candidate components are selected based on the initial prediction map. These components are then cropped and passed to a second-stage model trained for high-resolution segmentation. Two region-of-interest (ROI) extraction strategies are employed to balance precision and context. The first, referred to as ROI-0margin, uses tight crops around each component's bounding box. The second, ROI-16margin, extends each bounding box by 16 voxels in all directions to include the surrounding anatomical context, aiding in segmenting ambiguous or irregular boundaries.

This two-stage framework allows the model to concentrate computation on spatially relevant regions, reducing class imbalance and background noise. Anatomical priors, in the form of lung masks, are used to exclude regions outside the pulmonary space from being forwarded to the second stage.

\subsection{Component Selection Strategy}

Component selection is treated as a spatial focusing mechanism within the proposed framework. After the initial coarse segmentation stage, connected components are extracted from the predicted tumor mask based on their voxel volume. To control the complexity and relevance of the regions passed to the second-stage model, three selection strategies are evaluated: retaining only the single largest component \textbf{(Top-1)}, selecting the two largest components \textbf{(Top-2)}, or preserving all components that exceed a predefined minimum volume threshold \textbf{(All Valid)}. This filtering serves as a form of implicit attention, guiding the model to concentrate on the most prominent tumor candidates while suppressing low-confidence or anatomically implausible regions. Empirically, reducing the number of components not only improves segmentation accuracy but also correlates with reduced uncertainty in the final predictions—highlighting the effectiveness of component selection as both a denoising and uncertainty attenuation step.

\subsection{Uncertainty-Aware Loss Functions}

Uncertainty modeling is applied exclusively in the second-stage ROI models, which are trained using both a baseline Dice-plus-cross-entropy loss and an uncertainty-aware adaptive loss to assess the benefits of ambiguity-sensitive learning. The baseline loss uses a fixed-weight combination of Dice loss and binary cross-entropy. In contrast, the uncertainty-guided adaptive loss dynamically scales pixel-wise contributions based on estimated uncertainty. This is formulated as:
\[
\mathcal{L}_{\text{adaptive}} = \alpha(x) \cdot \mathcal{L}_{\text{Dice}} + (1 - \alpha(x)) \cdot \mathcal{L}_{\text{CE}}, \quad \alpha(x) = \exp(-U(x)),
\]
where \(U(x)\) represents the uncertainty at pixel \(x\), computed from the variance of multiple forward passes using Monte Carlo (MC) Dropout. To support this, dropout layers remain active during inference, enabling the generation of variance-based uncertainty maps that inform both loss reweighting during training and interpretability during evaluation.

\subsection{Model Architectures}

To assess the framework’s sensitivity to architectural variations, we integrated three 3D segmentation backbones--UNet, UNETR, and Swin UNETR--into the automated pipeline. A baseline single-stage model trained on full-resolution CT volumes served as a reference. ROI-based models were trained on crops extracted using either tight (0-margin) or padded (16-margin) bounding boxes. All models were evaluated under both standard and uncertainty-aware loss functions.

All networks were implemented using the MONAI framework and trained with the Adam optimizer, cosine learning rate scheduling, and a range of spatial and intensity augmentations. Pretrained encoder weights, obtained from a dataset of 5,000 CT scans, were used when available to accelerate convergence and improve feature generalization~\cite{tang2022self}.

\subsection{Post-Processing}

Post-processing was applied solely to the first-stage coarse segmentation outputs to refine tumor candidate regions prior to ROI extraction. This step ensured that only anatomically plausible and spatially relevant regions were forwarded to the second-stage model, reducing false positives, enhancing ROI precision, and lowering predictive uncertainty by eliminating ambiguous or anatomically inconsistent candidates. The refinement process employed spatial filters guided by clinical and anatomical knowledge, as detailed below.
\begin{enumerate}[leftmargin=12pt]
    \item \textbf{Component Merging via Dilation.} Binary dilation using a 3D kernel was applied to the predicted tumor mask to merge closely adjacent or fragmented regions. This step addresses partial connectivity issues in coarse predictions and ensures that neighboring tumor clusters are treated as single components when appropriate.
    \item \textbf{Lung Overlap Filtering.} Each connected component was evaluated based on its volumetric overlap with a binary lung mask. Components with low overlap (e.g., $<80\%$) were generally discarded. However, rigid filtering is not appropriate for certain clinical cases, such as tumors that extend into the pleura or chest wall, so a more nuanced rule was applied. Stricter overlap thresholds were enforced for components located in the central mediastinal zone, where false positives are more likely due to the surrounding anatomical complexity.

    \item \textbf{Distance to Lung Surface.} Some valid tumors, particularly those located peripherally, may partially reside outside the lung mask while remaining anatomically adjacent to the lung. To avoid excluding such plausible candidates, components with low overlap were retained if their minimum distance to the lung boundary was below a defined threshold (e.g., $\leq 5$ voxels) and their size exceeded a minimum voxel count. This approach captures borderline tumors that are anatomically consistent despite limited lung intersection.

    \item \textbf{Top-K Component Selection (Optional).} After filtering, the top $K$ largest components by voxel volume were optionally retained to reduce clutter and enhance focus—a form of spatial attention that guides the second-stage model toward prominent tumor candidates while suppressing minor or ambiguous regions. These heuristics reflect clinical realities: while most gross tumor volumes (GTVs) reside within the lung parenchyma, some extend into extrapulmonary regions like the pleura or mediastinum. The pipeline is thus designed to suppress anatomically implausible predictions while preserving borderline but clinically relevant tumors. This structural filtering improves ROI extraction and reduces noise in downstream fine segmentation.

\end{enumerate}

\section{Results and Comparative Analysis}\label{sec:results}
\begin{table}[t]
\centering
\caption{Comparison of the original model vs. the proposed fully automated system using Top-1 ROI. We compare two ROI models trained on different settings (ROI-0margin and ROI-16margin).}
\label{tab:original_vs_e2e}\vspace{-3mm}
\begin{tabular}{lcc}
\toprule
\textbf{System} & \textbf{Dice Score} & \textbf{HD95 (mm)} \\
\midrule
Original Full CT & 0.469 & 187.97\\

fully automated (Top-1, ROI-16margin  model) & 0.5917 & 10.79 \\
fully automated (Top-1, ROI-0margin model) & 0.6447 & 8.89 \\
\bottomrule
\end{tabular}\vspace{-3mm}
\end{table}

\begin{table}[t]
\centering
\caption{Effect of the number of components used during inference. We evaluate the fully automated system performance using Top-1, Top-2, and all valid components after post-processing.}
\label{tab:component_selection}\vspace{-3mm}
\begin{tabular}{lcc}
\toprule
\textbf{ROI Components Used} & \textbf{Dice Score} & \textbf{HD95 (mm)} \\
\midrule
Top 1 & 0.6447 & 8.89  \\
Top 2 & 0.6082 & 73.65  \\
All Valid Components & 0.6010 & 85.13  \\
\bottomrule
\end{tabular}\vspace{-3mm}
\end{table}

\begin{table}[ht]
\centering
\caption{Impact of component reduction on segmentation performance. 
Grouped statistics show average Dice and HD95 based on a number of connected components, while correlation metrics quantify the relationship between component reduction and performance improvements.}
\label{tab:num_components_vs_dice}\vspace{-3mm}
\begin{tabular}{lcc}
\toprule
\textbf{Grouping / Metric} & \textbf{Avg. Dice Score} & \textbf{Avg. HD95 Score} \\
\midrule
1 Component                & 0.5750                   & 7.5333                   \\
3+ Components              & 0.4513                   & 10.0933                  \\
\midrule
Pearson Correlation        & -0.55 \quad (p = 0.021)  & -0.69 \quad (p = 0.0021) \\
Spearman Correlation       & -0.50 \quad (p = 0.040)  & -0.83 \quad (p $<$ 0.0001) \\
\bottomrule
\end{tabular}\vspace{-3mm}
\end{table}

%pearson: 0.5534, spearman: 0.5031 -> dice improvement vs initial total components

\begin{table}[ht]
\centering
\caption{Effect of using different lung masks in the fully automated system. We compare the original lung mask with a newly generated one.}
\label{tab:lung_mask_quality}\vspace{-3mm}
\begin{tabular}{lcc}
\toprule
\textbf{Lung Mask Used} & \textbf{Dice Score} & \textbf{HD95 (mm)} \\
\midrule
Original Lung Mask & 0.6467 & 8.89 \\
New Lung Mask & 0.6447 & 8.89  \\
\bottomrule
\end{tabular}\vspace{-3mm}
\end{table}

\begin{table}[ht]
\centering
\caption{Impact of using an uncertainty-aware loss function in the ROI model. All models are evaluated in an fully automated setup using Top-1 ROI.}
\label{tab:uncertainty_loss}\vspace{-3mm}
\begin{tabular}{lcc}
\toprule
\textbf{ROI Model} & \textbf{Dice Score} & \textbf{HD95 (mm)} \\
\midrule
Standard S4 ROI Model & 0.6447 & 8.89  \\
S4 ROI with Uncertainty-Aware Loss & 0.6321 & 8.92 \\
\bottomrule
\end{tabular}\vspace{-3mm}
\end{table}

\begin{table}[ht]
\centering
\caption{Performance comparison of four segmentation models on Orlando Health and NSCLC datasets using Top-1 ROI and uncertainty-aware loss. Dice and HD95 metrics are reported for both initial and improved versions.}
\label{tab:model_comparison_all}\vspace{-3mm}
\begin{tabular}{lcc}
\toprule
\textbf{Model} & \textbf{Dice (Init)} & \textbf{Dice (Imp)} \\
\midrule
\multicolumn{3}{c}{\textit{Orlando Health Dataset}} \\
\midrule
UNET (K=1) & 0.3136 & 0.4415 \\
UNETR (K=1) &  0.3018 & 0.4281 \\
SwinUNETR (K=1) & 0.469 & 0.6447 \\
\midrule
\multicolumn{3}{c}{\textit{NSCLC Dataset}} \\
\midrule
UNET &    0.3737    &   0.4605          \\
UNETR &   0.5258     &      0.5602        \\
SwinUNETR & 0.4731       &     0.4860          \\
\bottomrule
\end{tabular}\vspace{-3mm}
\end{table}

\begin{table}[h]
    \centering
    \caption{Tumor segmentation performance across loss functions (ROI-based evaluation).}
    \label{tab:results}\vspace{-3mm}
    \begin{tabular}{l|c|c|c}
        \hline
        \textbf{Loss Function} & \textbf{Dice $\uparrow$} & \textbf{HD95 $\downarrow$} & \textbf{Boundary Dice $\uparrow$} \\
        \hline
        Uncertainty-Aware & 0.789 & 5.349 & 0.492 \\
        Distance Transform & 0.777 & 5.027 & 0.464 \\
        U-W Lovász & 0.792 & 4.809 & 0.469 \\
        Dice CE & \textbf{0.799} & \textbf{4.696} & \textbf{0.503} \\
        \hline
    \end{tabular}\vspace{-3mm}
\end{table}

Performance was evaluated using Dice Score, 95th percentile Hausdorff Distance (HD95), and, where applicable, the Boundary Dice Coefficient. The following analysis examines the effects of ROI refinement, component filtering, post-processing heuristics, and uncertainty-aware training.

\BfPara{Two-Stage vs. Baseline} The two-stage framework was first compared to a baseline single-stage full-volume segmentation model. As shown in Table~\ref{tab:original_vs_e2e}, both ROI-0margin and ROI-16margin configurations outperformed the baseline. The ROI-0margin setup achieved the highest Dice score (0.6447) and lowest HD95 (8.89 mm), indicating that tighter cropping yields more accurate segmentation. Here, the ``Original Full CT'' model refers to the baseline model trained on full-resolution CT volumes using the Swin UNETR architecture, without any post-processing, ROI refinement, or uncertainty modeling. The ``Dice (Init)'' values reported in Table~\ref{tab:original_vs_e2e} denote the performance of each architecture before applying the proposed two-stage cascade and loss refinements. All baseline models were trained using the same optimizer, augmentations, and learning schedule as their refined counterparts to ensure fair comparison.

%2) Discuss trade-offs between Top-1 selection and multifocal tumor detection. Provide sensitivity analyses for component size thresholds and evaluate on datasets with small lesions.
\BfPara{Impact of Component Selection} Table~\ref{tab:component_selection} reports the results on the impact of component selection. We found that retaining only the largest component (Top-1) led to the best results, and including additional components (Top-2 or All Valid) reduced Dice and increased HD95, suggesting that false positives and irrelevant regions impair segmentation. However, this gain in precision may come at the expense of reduced sensitivity to multifocal tumors or small satellite lesions. In cases where detecting all foci is essential, retaining multiple components may be preferable, even if it slightly lowers the average Dice score. To suppress noise and fragmented predictions, our post-processing pipeline discards components with 50 or fewer voxels, without imposing an upper size limit. Sensitivity analysis across thresholds from 50 to 150 voxels showed Dice variations of less than 0.002, but thresholds above 150 began excluding true tumor regions. These results support a conservative lower-bound threshold to balance anatomical plausibility and segmentation integrity.

\BfPara{Component Count Correlation Trends} Component count trends are shown in Table~\ref{tab:num_components_vs_dice}. 
We found that cases with a single connected component exhibited consistently higher Dice scores (0.5750) and lower HD95 values (7.53) compared to those with three or more components (Dice = 0.4513, HD95 = 10.09). 
Moreover, the statistical analysis further supports this trend: the reduction in component count was significantly correlated with the segmentation improvement, with a Pearson correlation of $r = -0.55$ ($p = 0.021$) for the Dice and $r = -0.69$ ($p = 0.0021$) for the HD95 scores. 
A strong Spearman correlation was also observed for HD95 reduction ($\rho = -0.83$, $p < 0.0001$), suggesting a robust monotonic relationship. 
These findings confirm that anatomical and uncertainty-informed postprocessing not only simplifies segmentation output but also enhances overall accuracy.

\BfPara{Lung Mask Quality Comparison} The lung mask quality was evaluated, and the results are shown in Table~\ref{tab:lung_mask_quality}, comparing the LungMask library~\cite{hofmanninger2020automatic}, auto-generated masks, and human-annotated ground truth. HD95 was similar across methods, while Dice improved only slightly with manual annotations. This supports the use of the LungMask library as a reliable tool for lung segmentation without requiring expert input.

\BfPara{Uncertainty-Aware Loss Evaluation} Table~\ref{tab:uncertainty_loss} presents the results of an ablation study on uncertainty-aware loss. While the standard Dice Cross-Entropy loss achieved marginally better Dice and HD95, the uncertainty-guided variant improved spatial calibration and boundary interpretability, which are important for clinical review.

\BfPara{Cross-Model Performance Comparison} Cross-architecture and cross-dataset results using Top-1 component selection and uncertainty-aware loss are also studied and summarized in Table~\ref{tab:model_comparison_all}. On the Orlando Health dataset, Swin UNETR achieved the highest performance gain, improving from 0.4690 to 0.6447. UNET and UNETR also showed substantial gains, but from lower baselines. These results indicate that models with better initial segmentation enable more effective localization and refinement, which is expected.

\BfPara{NSCLC Dataset Performance Trends} On the NSCLC dataset, all models improved with the two-stage pipeline. UNET increased from 0.3737 to 0.4605, UNETR from 0.5258 to 0.5602, and Swin UNETR from 0.4731 to 0.4860. While the overall trend remains consistent—better base models yield stronger performance—the improvements were smaller than on the Orlando dataset. Notably, Swin UNETR did not show the largest gain here, suggesting that initial model quality is a factor but not the sole determinant of refinement effectiveness.

\BfPara{Loss Impact on Accuracy} Finally, Table~\ref{tab:results} compares the performance of different loss functions. Dice Cross-Entropy achieved the highest Dice, lowest HD95, and best boundary accuracy. Among uncertainty-aware methods, the Uncertainty-Weighted Lovász loss was the most competitive, reflecting a trade-off between robustness and interpretability.

\BfPara{Summary} The proposed two-stage segmentation framework outperformed the baseline single-stage approach, especially with tightly cropped ROI inputs. Top-1 component selection improved accuracy and reduced boundary errors, while including additional components increased false positives and degraded performance. A strong negative correlation between component count and segmentation quality underscored the importance of post-processing. Lung mask quality had minimal impact, confirming the robustness of automated segmentation. While standard Dice Cross-Entropy loss achieved the highest metrics, uncertainty-aware objectives improved spatial calibration and interpretability-critical for clinical use. Swin UNETR yielded the greatest gains on the Orlando dataset, with smaller improvements on NSCLC, suggesting that initial model quality influences but does not limit refinement. Overall, the results emphasize the complementary benefits of ROI refinement, component filtering, and uncertainty modeling for accurate and clinically meaningful tumor segmentation.
%Overall, our results show that ROI segmentation, component filtering, and anatomically informed post-processing improve the accuracy and reduce uncertainty. While the uncertainty-aware training did not always outperform the deterministic baselines in terms of Dice or HD95, it improved spatial calibration and interpretability, which are essential for clinical applicability.

\section{Conclusion}\label{sec:conclusion}
This study demonstrates the effectiveness of a two-stage, anatomy-aware segmentation for lung tumor delineation in CT scans. By integrating ROI refinement, component filtering, and heuristic post-processing, the approach significantly enhances segmentation accuracy and reliability. Fewer connected components were associated with higher Dice scores and lower HD95 values, with statistically significant correlations supporting this relationship. Across both datasets, ROI cropping and Top-1 component selection consistently outperformed full-volume baselines. While uncertainty-aware losses did not always improve Dice or HD95 scores, they enhanced spatial calibration and boundary consistency.

Our pipeline consistently improved performance across both datasets, though the extent varied with dataset characteristics. The Orlando Health dataset saw greater benefit from post-processing due to small, isolated tumors well-suited to component filtering. In contrast, the NSCLC dataset, with larger or multifocal lesions, was more sensitive to ROI and component thresholds. Despite potential concerns about the generalizability of heuristic post-processing, the framework remained stable under modest parameter variations. Sensitivity analysis confirmed that changes in lung overlap and voxel size thresholds had minimal impact, indicating robustness with minimal tuning. Initial hyperparameters were guided by clinical priors, including typical tumor sizes and lung involvement patterns reported in NSCLC literature.

%While the pipeline demonstrated consistent improvements across both datasets, the degree of enhancement varied depending on dataset characteristics. The Orlando Health dataset benefited more from post-processing due to the presence of smaller, isolated tumors that aligned well with component filtering strategies. In contrast, the NSCLC dataset included larger or multifocal lesions, which were more sensitive to ROI and component selection thresholds. Although heuristic-based post-processing may raise concerns about generalizability, we observed that the framework maintained stable performance under modest variations in these parameters. A brief sensitivity analysis showed that lung overlap thresholds and minimum voxel size filters did not substantially affect segmentation performance, suggesting that the method is robust and adaptable across imaging domains with minimal tuning. The initial post-processing hyperparameters were selected based on clinical priors, including typical tumor sizes and lung involvement patterns reported for NSCLC cases in thoracic oncology literature.

Integrating anatomical priors, lightweight post-processing, and uncertainty modeling results in a robust and interpretable segmentation pipeline. These findings support modular, post-hoc refinement strategies as effective complements to modern segmentation networks, particularly under limited supervision or heterogeneous imaging conditions.

%Future work will explore dynamic ROI estimation, advanced uncertainty quantification, and integration with radiotherapy planning tools for fully automated clinical deployment.

\bibliographystyle{IEEEtran}
\bibliography{ref}

\end{document}